\documentclass[journal]{IEEEtran}
\usepackage{cite}
\usepackage{upgreek}
\usepackage{indentfirst}
\usepackage{amsmath}
\usepackage{graphicx}
\usepackage{subfigure}
\usepackage{algorithm}
\usepackage{algpseudocode}
\usepackage{multirow}
\usepackage{array}
\usepackage{flushend}
\usepackage{setspace}
\usepackage{float}
\usepackage{stfloats}
\usepackage{epstopdf}
\usepackage{subfigure}
\usepackage{multicol}
\usepackage{lscape}
\usepackage{bm}
\usepackage{pdflscape}

\begin{document}

\title{Dimensionality Reduction of Hyperspectral Imagery Based on Spatial-spectral Manifold Learning}

\author{Hong~Huang,~\IEEEmembership{Member,~IEEE,}~Guangyao~Shi,~Haibo~He,~\IEEEmembership{Fellow,~IEEE,}~Yule~Duan, and~Fulin~Luo
\thanks{This work was supported  by  the Basic and Frontier Research Programmes of Chongqing under Grant cstc2018jcyjAX0093, the Chongqing University Postgraduates Innovation Project under Grants  CYB18048 and CYS18035,  and the National Science Foundation of China under Grant 41371338.}
\thanks{H. Huang, G. Shi, and Y. Duan are with Key Laboratory of Optoelectronic Technology and Systems of the Education Ministry of China, Chongqing University, Chongqing 400044, China (e-mail: hhuang@cqu.edu.cn; shiguangyao@cqu.edu.cn; duanyule@cqu.edu.cn).}
\thanks{H. He is with the Department of Electrical, Computer, and Biomedical Engineering, University of Rhode Island, Kingston, RI 02881, USA (email:he@ele.uri.edu).}
\thanks{F. Luo is with State Key Laboratory of Information Engineering in Surveying, Mapping and Remote Sensing, Wuhan University, Wuhan 430079, China (e-mail: luoflyn@163.com).}
\thanks{Color versions of one or more of the figures in this paper are available online at http://ieeexplore.ieee.org.}
\thanks{Digital Object Identifier XXXX}}

\markboth{IEEE TRANSACTIONS ON CYBERNETICS,~Vol.~XX, No.~X, XX~XXXX}
{Huang \MakeLowercase{\textit{et al.}}: SSMRPE for Dimensionality Reduction of Hyperspectral Imagery}

\maketitle

\begin{abstract}
The graph embedding (GE) methods have been widely applied for dimensionality reduction of hyperspectral imagery (HSI). However, a major challenge of GE is how to choose proper neighbors for graph construction and explore the spatial information of HSI data. In this paper, we proposed an unsupervised dimensionality reduction algorithm termed spatial-spectral manifold reconstruction preserving embedding (SSMRPE) for HSI classification. At first, a weighted mean filter (WMF) is employed to preprocess the image, which aims to reduce the influence of background noise. According to the spatial consistency property of HSI, the SSMRPE method utilizes a new spatial-spectral combined distance (SSCD) to fuse the spatial structure and spectral information for selecting effective spatial-spectral neighbors of HSI pixels. Then, it explores the spatial relationship between each point and its neighbors to adjusts the reconstruction weights for  improving the efficiency of manifold reconstruction. As a result, the proposed method can extract the discriminant features and subsequently improve the classification performance of HSI. The experimental results on PaviaU and Salinas hyperspectral datasets indicate that SSMRPE can achieve better classification accuracies in comparison with some state-of-the-art methods.
\end{abstract}


\IEEEpeerreviewmaketitle

\section{Introduction}
\label{sec:Introduction}

\IEEEPARstart{B}y observing digital images in hundreds of continuous narrow spectral bands spanning the visible to infrared wavelengths, hyperspectral imagery (HSI) contains both detailed spatial structure and spectral information \cite{Jia20181176, Zhai20181, Zhang201816}. Due to its ability to distinguish more subtle differences between ground cover classes than traditional multi-spectral imagery \cite{Peng20181, Han2017195}, HSI has been widely used in many fields such as environmental monitoring, precision agriculture, urban planning, and earth observation \cite{Wang20163912, Jia20163174, Guo2017347}. Classification of each pixel in HSI plays a crucial role in these real applications, but the traditional classification methods commonly cause the Hughes phenomena because of the high dimensional characteristic of spectral features \cite{Luo20181, Pu20147008}. Therefore, the most important and urgent issue is how to reduce the number of bands largely with some valuable intrinsic information preserved \cite{Yang20181, Luo2017790, Pan20176085}.

In recent years, many dimensionality reduction (DR) methods have been proposed to reduce the number of spectral bands in HSI. Principal component analysis (PCA) \cite{Datta2017850} and linear discriminant analysis (LDA) \cite{Luo20151} are two widely used subspace learning methods which project high-dimensional data into a lower-dimensional embedding space by using a set of optimal basis vectors, but they cannot reveal the intrinsic structure in HSI data \cite{Gan201725069, Yuan2017934}. While manifold learning methods are useful to analyze the data that lie on or near a manifold in the original data space, many manifold learning methods have been introduced for discovering the intrinsic structure in high-dimensional data \cite{Huang20155160, Sun2017506, Zhou20161667}, such as local linear embedding (LLE) \cite{Zhang201625}, isometric feature mapping (ISOMAP) \cite{Li20171532}, laplacian eigenmaps (LE) \cite{Doradomunoz201683}, and t-distributed stochastic neighbor embedding (t-SNE) \cite{Kruiger2017283}. However, these methods are non-linear techniques, and the issue of how to map unknown data points into embedding space remains difficult \cite{Fang20171712, Yan2015849, Crawford2011207}. To solve this problem, many linear manifold learning methods were developed to directly map unknown samples into embedding space, e.g. locality preserving projection (LPP) \cite{Deng2018277}, neighborhood preserving embedding (NPE) \cite{Lv20174} and parametric supervised t-SNE \cite{Yu2017849}. In order to unify these methods, a graph embedding (GE) framework has been proposed to analyze the DR methods on the basis of statistics or geometry theory \cite{Luo20174389, Tan201519, Wen20164272}.

However, the above DR methods only consider the spectral information and neglect the spatial correlations among pixels in HSI, which restricts their discriminant capability for classification in real applications \cite{Xia20164971, Xia20152532, Yue2015468, Jia20152473}. Therefore, many spatial-spectral DR methods have been proposed to fuse spatial correlation and spectral information for enhancing the classification performance. Wei \emph{et al}. \cite{Wei20121249} proposed a spatial coherence-neighborhood preserving embedding (SC-NPE) method, which considered spatial context of pixels by adopting the difference between the surrounding patch of pixels, and then mapped the raw data into the low-dimensional space through an optimized local linear embedding. Zhou \emph{et al}. \cite{Zhou20151082} developed a spatial-domain local pixel neighborhood preserving embedding (LPNPE) method, and it seeks a linear projection matrix such that the local pixel neighborhood preserving scatter is minimized and the total scatter is maximized in the projected space simultaneously. Feng \emph{et al}. \cite{Feng2015224} defined discriminate spectral-spatial margins (DSSMs) to reveal the local information of hyperspectral pixels and explore the global structure of both labeled and unlabeled data via low-rank representation. The aforementioned spatial-spectral combined methods just use the spatial information to represent the similarity relationship or reveal the spatial neighborhood relationship of HSI data within a certain spatial window, which ignore the influence of spatial information in the construction of adjacency graph.

To overcome the aforementioned drawbacks, we proposed a new unsupervised DR method termed spatial-spectral manifold reconstruction preserving embedding (SSMRPE) for hyperspectral imagery classification. The SSMRPE method makes full use of spatial structure and spectral information in HSI to extract discriminant features for classification, and the main characteristics of this DR method can be concluded as: 1) As a preprocessing step, a spatial weighted mean filter (WMF) method is explored to reduce noise and smoothen the homogeneous regions in HSI; 2) Compared with traditional Euclidean distance, the proposed spatial-spectral combined distance (SSCD) is a helpful way to choose effective spatial-spectral neighbors by incorporating the spatial structure and spectral information simultaneously; 3) A spatial-spectral adjacency graph is constructed to discover the manifold structure of HSI data, and the aggregation of data is enhanced through adjusting the reconstruction weight of spatial neighbors to extract the discriminant features. Experimental results on PaviaU and Salinas hyperspectral datasets show that the proposed SSMRPE method achieved better classification performance than some state-of-art DR methods.

This paper is organized as follows. In Section II, we briefly review some related works. Section III details our proposed method. Experimental results on two real hyperspectral data sets are presented in Section IV to demonstrate the effectiveness of the proposed SSMRPE method. Finally, Section V provides some concluding remarks and suggestions for future work.

\section{Related works}
\label{sec:Related Works}
Suppose that a HSI dataset consists of \emph{D} bands, each pixel can be denoted as a vector ${\bf{x}}_{i}\in {\bf{R}}^{D}$($\emph{i}=1,2,\ldots,n$), where \emph{n} refers to the number of HSI pixels. The class label of $\ell({{\bf{x}}_{i}}) \in \{ 1, 2, \cdots, c \}$, where \emph{c} is the number of land cover types. The goal of dimensionality reduction is to map ${\bf{X}}\in {\bf{R}}^{D}$ to ${\bf{Y}}\in {\bf{R}}^{d}$, where $D \gg d$. For the linear DR methods, ${\bf{Y}} = [{{\bf{y}}_{1}}, {{\bf{y}}_{2}}, {{\bf{y}}_{3}}, \cdots, {{\bf{y}}_{n}}] \in {{\bf{R}}^{d \times n}}$ is replaced by ${\bf{Y}} = {{\bf{A}}^T}{\bf{X}}$, where ${\bf{A}} \in {{\bf{R}}^{D \times d}}$ is the corresponding projection matrix.

\subsection{Weighted Mean Filtering (WMF)}
To reduce noise and smoothen the homogeneous regions in the HSI, a spatial WMF is used to preprocess the pixels. Assuming that the coordinate of ${\bf{x}}_{i}$ is denoted as $(p_{i}, q_{i})$, then the adjacent space centered at ${\bf{x}}_{i}$ can be defined as
\begin{equation}
\Omega({\bf{x}}_{i})=\left \{ {\bf{x}}_{i}(p,q)|p \in [p_{i}-t, p_{i}+t],q \in [q_{i}-t,q_{i}+t]  \right \}
\end{equation}
where $t = {{(w - 1)} \mathord{\left/{\vphantom {{(w - 1)} 2}} \right.\kern-\nulldelimiterspace} 2}$, $w$ is a positive odd number, and it indicates the size of spatial window. Denoting the pixels in the adjacent space $ \Omega({\bf{x}}_{i})$ as $ {\bf{x}}_{i}, {\bf{x}}_{i1}, {\bf{x}}_{i2},\cdots, {\bf{x}}_{i(w^2-1)}$, Then, the filtered pixel ${\bf{x}}_{i}^{\prime}$ can be represented with a weighted summation as follows:
\begin{equation}
{\bf{x}}_{i}^{\prime} =\frac{{\sum\nolimits_{{\bf{x}}_{j} \in\Omega ({\bf{x}}_{i})} {\emph{v}_{j}{\bf{x}}_{j}} }}{{\sum\nolimits_{{\bf{x}}_{j} \in \Omega ({\bf{x}}_{i})} {\emph{v}_{j}} }}= \frac{{{\bf{x}}_{i} + \sum\nolimits_{k = 1}^{{w^2} - 1} {{\emph{v}_{k}}{\bf{x}}_{ik}} }}{{1 + \sum\nolimits_{k = 1}^{{w^2} - 1} {{\emph{v}_{k}}}}}
\end{equation}
where ${\emph{v}_{k}}=\textrm{exp}\{-\gamma_{0}\|{\bf{x}}_{i}-{\bf{x}}_{ik}\|^{2}\}$ is the weight of ${\bf{x}}_{ik}$, and $\gamma _{0}$ is a constant which is empirically set 0.2 in the experiments. The WMF method adjusts the degree of filtering by spatial window \emph{w}, and it can effectively decrease the influence of noise for enhancing the spectral similarity between pixels from the same class.

\subsection{Neighborhood Preserving Embedding (NPE)}
NPE can be regarded as a linear approximation to locally linear embedding (LLE), and it can directly map unknown samples into embedding space where the local manifold structure of data can be preserved. The outline of NPE can be summarized as follows:
\\1) Construct an adjacency graph ${\bf{G}}$. ${\bf{G}}$ is composed of \emph{n} nodes, and \emph{i}-th node corresponds to sample ${\bf{x}}_{i}$. Node \emph{i} and \emph{j} are connected by an edge if ${\bf{x}}_{j}$ is among \emph{k} nearest neighbors of ${\bf{x}}_{i}$. The common ways of selecting neighbors are $\emph{k-}$nearest neighbors and ${\bf{\varepsilon\emph{-}}}$neighborhood.
\\2) Compute the weight matrix ${\bf{W}}$. Let $w_{ij}$ denote the weight of ${\bf{x}}_{i}$ and ${\bf{x}}_{j}$, and $w_{ij}$ = 0 if there is no such edge between them; otherwise $w_{ij}$ can be calculated by minimizing the following reconstruction error function:
\begin{equation}
\min \sum\limits_{i = 1}^N {||{\bf{x}}_{i} - \sum\limits_{j = 1}^k {{w_{ij}}{\bf{x}}_{j}} |{|^2}}  = \min \sum\limits_{i = 1}^N {||\sum\limits_{j = 1}^k {{w_{ij}}({\bf{x}}_{i} - } {\bf{x}}_{j})|{|^2}}
\end{equation}
with constrains $\sum\limits_j {{w_{ij}} = 1}, j = 1, 2, \cdots, \emph{k}$
\\3) Calculate the projection matrix ${\bf{A}}$. To preserve the local manifold structure on high-dimensional data, NPE assumes that the low-dimensional embedding ${\bf{y}}_{i}$ can be approximated by the linear combination of its corresponding neighbors. Therefore, a reasonable criterion for choosing a good project matrix ${\bf{A}}$ is to minimize the objective function as
\begin{equation}
\left\{\begin{array}{l}\min\sum\limits_{i = 1}^N {|{\bf{y}}_{i} - \sum\limits_{j = 1}^k {{w_{ij}}{\bf{y}}_{ij}} {|^2}}=\min{{\bf{A}}^T}{\bf{X}}{\bf{M}}{{\bf{X}}^T}{\bf{A}}\\ \textrm{s.t.}\sum\limits_{i = 1}^N {{\bf{y}}_{i} = 0,\frac{1}{N}} {\bf{A}}{{\bf{A}}^T} = {\bf{I}}
\end{array} \right.
\end{equation}
where ${\bf{M}} = ({\bf{I}} - {\bf{W}})({\bf{I}} - {\bf{W}})^T $ and ${\bf{I}} = \textrm{diag} [1, \cdots,1]$.

\section{PROPOSED METHOD}

To effectively reveal the intrinsic manifold structure of hyperspectral data, a spatial-spectral manifold reconstruction preserving embedding (SSMRPE) method was proposed for DR of HSI. SSMRPE chooses spatial-spectral neighbors by incorporating the spatial structure and spectral information, which is more effective to choose the proper neighbors from HSI pixels, especially for spectrally-similar pixels from different classes. To further enhance the discriminating power of feature learning, it exploits the spatial-spectral neighbors to construct a spatial-spectral adjacency graph, and adjusts the reconstruction weight of them by their spatial coordinates in HSI. Thus, SSMRPE makes full use of spatial-spectral combined information of HSI data, and it can obtain discriminating features to improve the classification performance.The flowchart of the proposed algorithm is shown in Fig. 1.
\begin{figure*}[!htbp]
\centering
\includegraphics[scale=0.42]{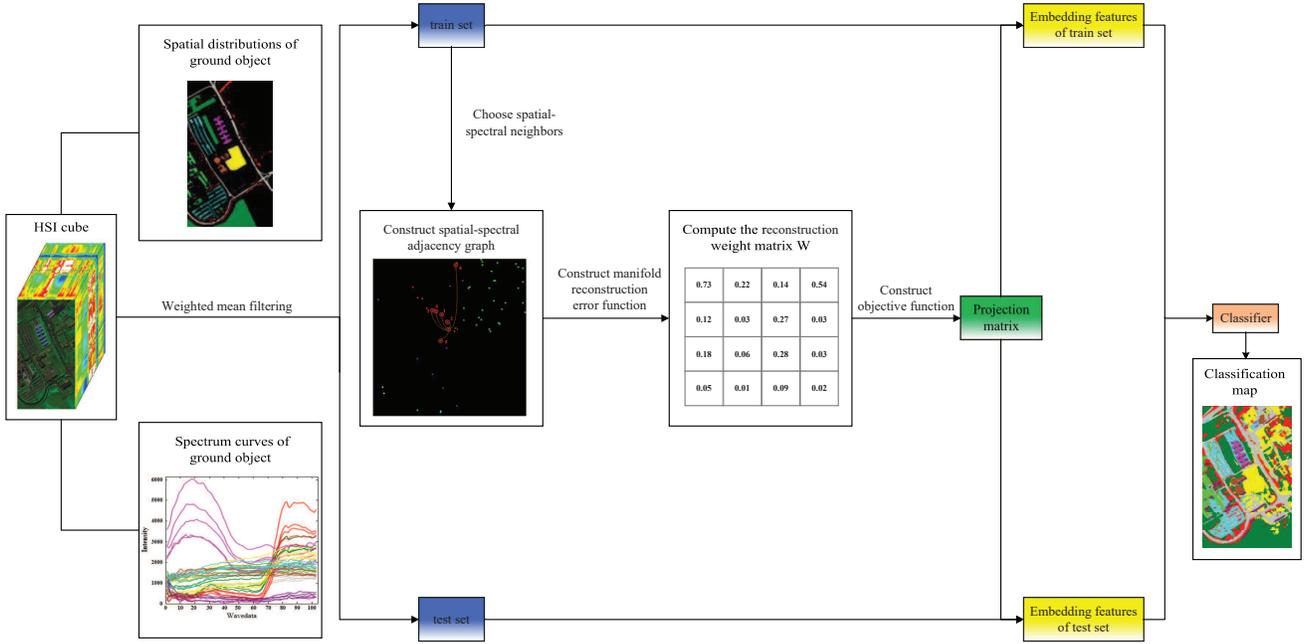}
\caption{Flowchart of the proposed SSMRPE method.}
\label{SSMRPE_fig}
\end{figure*}

In hyperspectral image, the pixels are usually spatially related, the pixels within a small neighborhood usually possess the spatial distribution consistency of ground objects. Therefore, neighborhood pixels can be utilized to fuse spatial and spectral information for measuring similarity.

In order to explore the spatial-spectral combined information in HSI, a spatial-spectral combined distance (SSCD) is proposed to choose effective neighbors for the construction of adjacency graph, which explores the adjacent space $\Omega({\bf{x}}_{i})$ to measure the similarity between data points. The overview of SSCD is shown in Fig. 2.
\begin{figure*}[!htbp]
\centering
\includegraphics[scale=0.62]{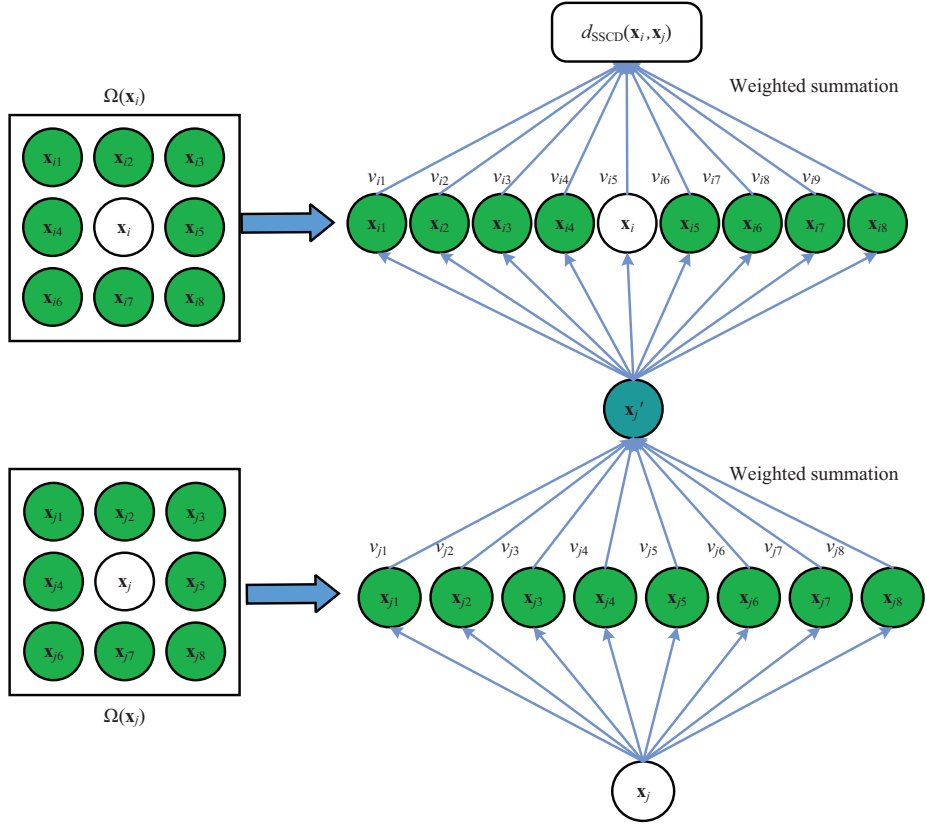}
\caption{Overview of spatial-spectral combined distance (SSCD). (Note that the green circles represent the spatial neighbors of ${\bf{x}}_{i}$ and ${\bf{x}}_{j}$, respectively.)}
\label{SSCD_fig}
\end{figure*}
For pixels ${\bf{x}}_{i}$ and ${\bf{x}}_{j}$ in HSI, ${\bf{x}}_{i}^{\prime}$ and ${\bf{x}}_{j}^{\prime}$ are the corresponding filtered pixels by using the weighted mean filter (WMF) method. Suppose the adjacent spaces of ${\bf{x}}_{i}$ and ${\bf{x}}_{j}$ are $\Omega({\bf{x}}_{i})$ and $\Omega({\bf{x}}_{j})$, respectively, the spatial-spectral combined distance (SSCD) can be defined as
\begin{equation}
{d_{\textrm{SSCD}}}({\bf{x}}_{i},{\bf{x}}_{j}) = d(\Omega ({\bf{x}}_{i}),{\bf{x}}_{j}^{\prime})
\end{equation}
where $d(\Omega ({\bf{x}}_{i}),{\bf{x}}_{j}^{\prime})$ is the distance between $\Omega ({\bf{x}}_{i})$ and ${\bf{x}}_{j}^{\prime}$, it is defined as follows:
\begin{equation}
d({\bf{x}}_{j}^{\prime},\Omega ({\bf{x}}_{i})) = \frac{{\sum\limits_{s = 1}^{{w^2}} {{\emph{v}_{is}}||{{\bf{x}}_{j}^{\prime}} - {\bf{x}}_{is}||} }}{{\sum\limits_{s = 1}^{{w^2}} {{\emph{v}_{is}}} }}{\rm{}}{\kern 1pt} {\kern 1pt} {\kern 1pt} {\kern 1pt} {\kern 1pt} {\kern 1pt} {\bf{x}}_{is}\in \Omega ({\bf{x}}_{i})
\end{equation}
in which ${\emph{v}_{is}}$ is the weight of ${\bf{x}}_{is}$, and it can be calculated by a heat kernel function as
\begin{equation}
{\emph{v}_{is}} = \exp \{ {{ - ||{\bf{x}}_{j}^{\prime} - {\bf{x}}_{is}|{|^2}} \mathord{\left/
 {\vphantom {{ - ||{\bf{x}}_{j}^{\prime} - {\bf{x}}_{is}|{|^2}} {\sigma _j^2}}} \right.
 \kern-\nulldelimiterspace} {\sigma _j^2}}\}
\end{equation}
where ${\sigma _j}$ is set to be the average value of $||{\bf{x}}_{j}^{\prime} - {\bf{x}}_{is}||$, $s = 1,2,\cdots,{w^2}$, that is
\begin{equation}
{\sigma_j} = \frac{1}{{{w^2}}}\sum\limits_{s = 1}^{{w^2}} {||{\bf{x}}_{j}^{\prime} - {\bf{x}}_{is}||}
\end{equation}

With the modified distance by SSCD, a number of spatial-spectral neighbors can be obtained for adjacency graph. To further illustrate the effectiveness of SSCD, we crop an image block from the PaviaU hyperspectral image, as shown in Fig.3 (a). Then we randomly select some pixels from the image block as training samples, and the distributions of all training samples are shown in Fig.3 (b). ${\bf{x}}_{i}$ is set as the target pixel that is denoted with a brown circle, and then its spectral neighbors, spatial neighbors and spatial-spectral neighbors are chosen, respectively. Note that the window size of spatial-based methods is $9 \times 9$. All the neighbors are indicated by red circles, which are connected with ${\bf{x}}_{i}$ by brown lines. The distributions for three different types of neighbors are shown in Fig.3 (c-e), and their corresponding spectral curves of selected neighbors are shown in Fig.3 (f-h).
\begin{figure*}[!htbp]
\centering
\includegraphics[scale=0.95]{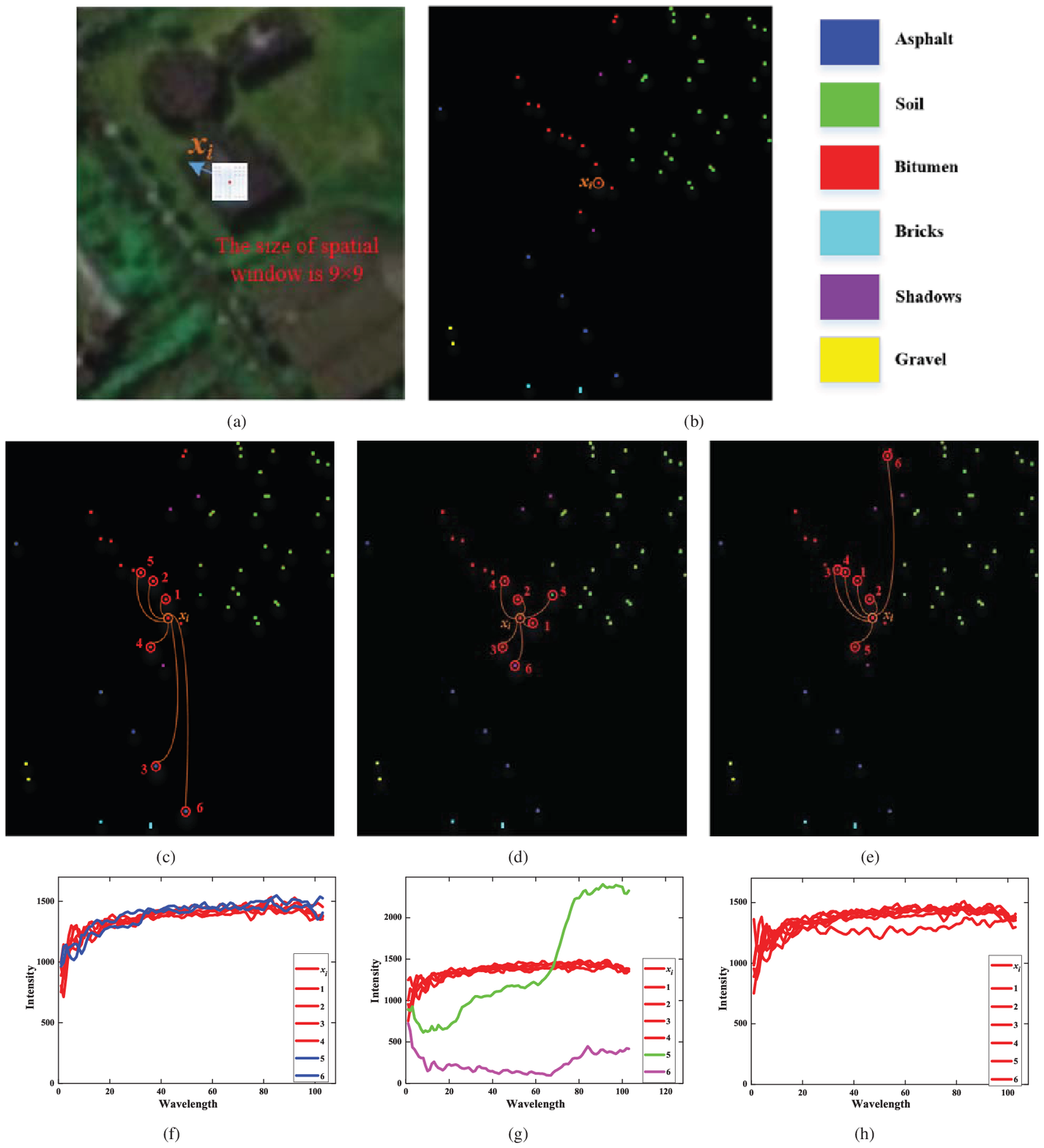}
\caption{Diagrams of spectral, spatial and spatial-spectral neighbors. (a)  Original image (b) Randomly selected training samples (c) Spectral neighbors (d) Spatial neighbors (e) Spatial-spectral neighbors (f) Curves of spectral neighbors (g) Curves of spatial neighbors (h) Curves of spatial-spectral neighbors}
\label{tu8}
\end{figure*}

As we can see from Fig.3 (a) and (f), spectral neighbors measure the similarity of pixels only based on spectral features while neglect the spatial distributions of ground objects, and the selected neighbors often contain pixels with similar spectral curves which come from different classes. From Fig.3 (d) and (g), spatial neighbors consider the spatial relationship between pixels, but the spatial neighbors may include pixels come from different classes, especially at land cover boundaries, which will produce negative influence on DR of HSI. As in Fig.3 (e) and (h), the spatial-spectral combined information between pixels are explored to find effective neighbors, which is beneficial to construct the adjacency graph for DR.

After obtaining the spatial-spectral neighbors, a spatial-spectral adjacency graph ${\bf{G}}^{ss} = \{{\bf{X}},{\bf{W}}^{ss}\}$ can be constructed, where ${\bf{X}}$ is the vertex of the graph and ${\bf{W}}^{ss}$ is the weight matrix. In graph ${\bf{G}}^{ss}$, if two pixels are neighbors, an edge is connected between them, otherwise they should not be connected. In general, the selected neighbors may have different effect in the reconstruction process due to their spatial relationships. In hyperspectral image, if a neighbor pixel is closer to the target pixel, there would be a high probability of them being in the same class, and the weight between them should be bigger in the reconstruction process. Therefore, a spatial coordinate distance $d_{\textrm{SCD}}$ is introduced to denote the spatial relationship.
\par Denoting the coordinates of ${\bf{x}}_{i}$ and ${\bf{x}}_{j}$ in HSI as $(p_{i}, q_{i})$ and $(p_{j}, q_{j})$, then the spatial coordinate distance $d^{ij}_{\textrm{SCD}}$ is given by the Euclidean distance between their coordinates as
\begin{equation}
d_{\textrm{SCD}}^{ij} = d_{\textrm{SCD}}({\bf{x}}_{i},{\bf{x}}_{j}) = \sqrt {{{({p_{i}} - {p_{j}})}^2}{\rm{ + (}}{q_{i}} - {q_{j}}{{\rm{)}}^2}}
\end{equation}

To preserve the spatial relationship of HSI data, the reconstruction error function for the optimal weights is defined as follows:
\begin{equation}
\left\{ {\begin{array}{*{20}{c}}
{\min \sum\limits_{i = 1}^N {||\sum\limits_{j = 1}^k {w_{ij}^{\prime}(\frac{{{\bf{x}}_{i} - {\bf{x}}_{j}}}{{d_{\textrm{SCD}}^{ij}}})} |{|^2}} }\\
{\textmd{s.t.}{w_{ij}^{\prime}} = 0{\kern 1pt} {\kern 1pt} {\kern 1pt} {\kern 1pt} {\kern 1pt} {\kern 1pt} {\kern 1pt} {\kern 1pt} {\kern 1pt} {\kern 1pt} {\kern 1pt} {\kern 1pt} {\kern 1pt} {\kern 1pt} \forall {\kern 1pt} {\kern 1pt} {\kern 1pt}{\bf{x}}_{j}\notin k({\bf{x}}_{i})}
\end{array}} \right.
\end{equation}
In Eq.(10), $({\bf{x}}_{i} - {\bf{x}}_{j})/d^{ij}_{\textrm{SCD}}$ indicates the spatial-spectral combined measure between ${\bf{x}}_{i}$ and ${\bf{x}}_{j}$.

Suppose that ${\bf{x}}_{i}^{k}$ represents the \emph{k}-th spatial-spectral neighbor of ${\bf{x}}_{i}$, ${\bf{h}}_{i}^{k}=({\bf{x}}_{i} - {\bf{x}}_{i}^{k})/d^{ik}_{\textrm{SCD}}$ is the spatial-spectral combined measure between ${\bf{x}}_{i}$ and its \emph{k}-th spatial-spectral neighbor, and \emph{k} is the number of spatial-spectral neighbors, then the objective function can be simplified into
\begin{equation}
\begin{array}{l}
\min \sum\limits_{i = 1}^N {||\sum\limits_{j = 1}^k {{{w}_{ij}^{\prime}}(\frac{{{\bf{x}}_{i} - {\bf{x}}_{j}}}{{d_{\textrm{SCD}}^{ij}}})} |{|^2}} = \min \sum\limits_{i = 1}^N {{\bf{w}}_i^{\prime T}{{\bf{z}}_{i}}} {{\bf{w}}_i^{\prime}}
\end{array}
\end{equation}
where ${{\bf{z}}_i} = {[h_i^1,h_i^2,h_i^3,...,h_i^k]^T}[h_i^1,h_i^2,h_i^3,...,h_i^k]$ and ${{\bf{w}}_i^{\prime}} = [{w_{i1}^{\prime}},{w_{i2}^{\prime}},{w_{i3}^{\prime}},\cdots,{w_{ik}^{\prime}}]$.
Then, the objective function can be expressed as the following optimization problem:
 \begin{equation}
\left\{ {\begin{array}{*{20}{c}}
{\min \sum\limits_{i = 1}^N {{\bf{w}}^{\prime T}_i{z_i}{{\bf{w}}_i^{\prime}}}}\\
{\textmd{s.t.}\sum\limits_{j = 1}^k {{{w}_{ij}^{\prime}} = 1} }
\end{array}} \right.
\end{equation}
With the Lagrange multiplier method, $w^{\prime}_{ij}$ is given as follows:
\begin{equation}
{w^{\prime}_{ij}} = \frac{{\sum\limits_{m = 1}^k {{{(z_i^{jm})}^{ - 1}}} }}{{\sum\limits_{p = 1}^k {\sum\limits_{q = 1}^k {{{(z_i^{pq})}^{ - 1}}} } }}
\end{equation}
where $ z_i^{jm} = {(h_i^j)^T}h_i^m $ and $ z_i^{pq} = {(h_i^p)^T}h_i^q $.

After obtained the weight matrix ${\bf{W}}^{\prime}$, the projection vector {\bf{A}} for low-dimensional embedding can be achieved by solving the following optimization problem:
\begin{equation}
\left\{ \begin{array}{l}
\min {\kern 1pt} \sum\limits_{i = 1}^N {|{{\bf{y}}_i} - \sum\limits_{j = 1}^k {{w_{ij}^{\prime}}{{\bf{y}}_j}} {|^2}} \\
\textmd{s.t.}\sum\limits_{i = 1}^N {{{\bf{y}}_i} = 0,{\kern 1pt} {\kern 1pt} {\kern 1pt} \frac{1}{N}} {\bf{A}}{{\bf{A}}^T} = {\bf{I}}
\end{array} \right.
\end{equation}

With some mathematical operations, (14) can be reduced as
\begin{equation}
\begin{array}{l}
\min \sum\limits_{i = 1}^N {||{{\bf{y}}_i} - \sum\limits_{j = 1}^k {{w_{ij}^{\prime}}{{\bf{y}}_j}} |{|^2}} \\
 = \min \sum\limits_{i = 1}^N {||{\kern 1pt} \sum\limits_{j = 1}^k {{w_{ij}^{\prime}}({{\bf{y}}_i}} {\kern 1pt} {\kern 1pt}  - {{\bf{y}}_j})|{|^2}} \\
\vspace{1.1ex}
 = \min \sum\limits_{i = 1}^N {||{\kern 1pt} {\bf{Y}}{{\bf{I}}_i} - {\bf{Y}}{{\bf{w}}_i^{\prime}}|{|^2}} \\
\vspace{1.4ex}
 \displaystyle = \min tr({\bf{Y}}({\bf{I}} - {\bf{W}}^{\prime}){({\bf{I}} - {\bf{W}}^{\prime})^T}{{\bf{Y}}^T})\\
  \displaystyle= \min {{\bf{A}}^T}{\bf{XM}}'{{\bf{X}}^T}{\bf{A}}
\end{array}
\end{equation}
in which ${\bf{M}}^{\prime} = ({\bf{I}} - {\bf{W}}^{\prime}) ({\bf{I}} - {\bf{W}}^{\prime})^T $and \emph{I} = \textmd{diag} [1, $\cdots$ ,1]. (15) can be solved by Lagrange multiplier, and it can be transformed into the following form:
\begin{equation}
{\bf{XM}}^{\prime}{{\bf{X}}^T}{\bf{A}} = \lambda {\bf{X}}{{\bf{X}}^T}{\bf{A}} \Rightarrow {({\bf{X}}{{\bf{X}}^T})^{ - 1}}{\bf{X}}{\bf{M}}^{\prime}{{\bf{X}}^T}{\bf{A}} = \lambda {\bf{A}}
\end{equation}
where $\lambda$ is the eigenvalue of (16). With the eigenvectors $a_1$, $a_2$, $\cdots$ , $a_d$ corresponding to the first \emph{d} eigenvalues, the optimal projection matrix can be represented as {\bf{A}} = [$a_1$ $a_2$ $\ldots$ $a_d$]. Then the embedding of high-dimensional data in the low-dimensional space can be denoted as ${\bf{Y}} = {\bf{A}}^T{\bf{X}}$.

In summary, the proposed method makes full use of spatial-spectral combined information of HSI data, and it can obtain discriminating features to improve the classification performance.
The detailed process of SSMRPE method is given in Algorithm 1.
\begin{algorithm*}[!htbp]
\renewcommand{\algorithmicrequire}{\textbf{Input:}}
\renewcommand{\algorithmicensure}{\textbf{Output:}}
\caption{SSMRPE}
\label{alg_SSMRPE}
\begin{algorithmic}[1]
\Require
\emph{D}-dimensional data set $X = [{\bf{x}}_{1},{\bf{x}}_{2},{\bf{x}}_{3},\cdots,{\bf{x}}_{n}] \in{{\bf{R}}^{D \times n}}$ and their corresponding class labels $\{ {l_1},{l_2},\cdots,{l_n}\}$, embedding dimension $\emph{d}(\emph{d} \ll \emph{D})$, window size $\emph{w} > 0$, neighbor number $\emph{k} > 0$ and  $\gamma_{0}=0.2$.
\State for \emph{i} = 1 to \emph{n} do
\State ~~~~find neighbor pixels of ${\bf{x}}_{i}$ by (1).
\State ~~~~for \emph{s} = 1 to do
\State ~~~~~Compute the weights of neighbor pixels:${\emph{v}_{k}}=\textrm{exp}\{-\gamma_{0}\|{\bf{x}}_{i}-{\bf{x}}_{ik}\|^{2}\}$
\State ~~~~end for
\State ~~~~Obtain the filtered pixel ${\bf{x}}_{i}^{\prime} =\frac{{\sum\nolimits_{{\bf{x}}_{j} \in\Omega ({\bf{x}}_{i})} {\emph{v}_{j}{\bf{x}}_{j}} }}{{\sum\nolimits_{{\bf{x}}_{j} \in \Omega ({\bf{x}}_{i})} {\emph{v}_{j}} }}$
\State end for
\State ${\bf{X}}^{\prime} = [{{\bf{x}}_1^{\prime}},{{\bf{x}}_2^{\prime}},{{\bf{x}}_3^{\prime}}, \cdots,{{\bf{x}}_n^{\prime}}]$
\State for \emph{i} = 1 to \emph{n} do
\State ~~~~Compute the spatial-spectral combined distance as (5).
\State ~~~~find \emph{k-}nearest spatial-spectral neighbors $[{\bf{x}}_{i}^1,{\bf{x}}_{i}^2,{\bf{x}}_{i}^3,\cdots,{\bf{x}}_{i}^k]$ of ${\bf{x}}_{i}$
\State ~~~~compute spatial coordinate distance $d_{\textmd{SCD}}^{ij}$ between ${\bf{x}}_{i}$ and its spatial-spectral neighbors.
\State end for
\State Compute the reconstruction weight of ${\bf{x}}_{i}$ by (11).
\State Solve the generalized eigenvalue problem as (16).
\State Obtain the projection matrix with the \emph{d} smallest eigenvalues corresponding eigenvectors: ${\bf{A}} = [a_1,a_2, \cdots,a_d]\in {{\bf{R}} ^{D \times d}}$
\Ensure
${\bf{Y}} = {{\bf{A}}^T}{\bf{X}} \in {{\bf{R}} ^{d \times n}}$
\end{algorithmic}
\end{algorithm*}

\section{EXPERIMENTAL RESULTS AND ANALYSIS}
In this section, the Pavia University and Salinas Valley hyperspectral data sets were employed to demonstrate the effectiveness of the proposed SSMRPE method, and it was compared with some state-of-art DR algorithms.
\subsection{Data Description}
1) Pavia University data set: This data set is a hyperspectral image of the Pavia University acquired by the ROSIS sensor in 2002. The area possesses a spatial size of $610 \times 340$ pixels, and the uncorrected data contains 115 spectral bands ranging from 0.43 to 0.86 $\upmu$m with a spatial resolution of 1.3 m. The corrected data has 103 bands after the 12 bands with serious water absorption are removed. The data set contains nine land cover types. The HSI in false color and its corresponding ground truth are shown in Fig. 4.
\begin{figure}[!htbp]
\centering
\includegraphics[scale=1.12]{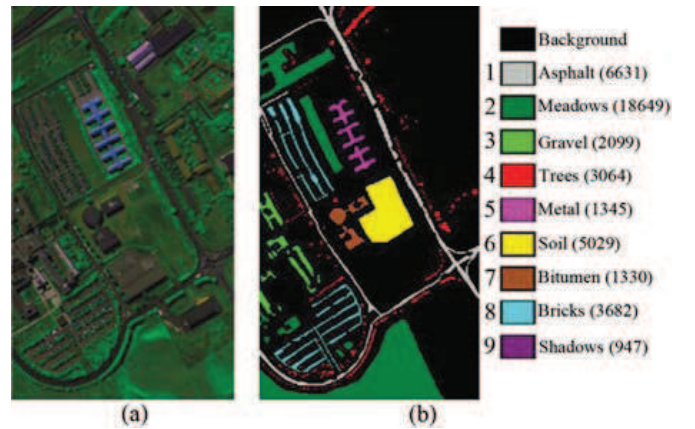}
\caption{PaviaU hyperspectral remote sensing image. (a) HSI in false color. (b) Corresponding ground truth. (Note that the number of samples for each class is shown in brackets.)}
\label{PaviaU_gt_fig}
\end{figure}

\par 2) Salinas data set: This HSI data set was collected by an airborne visible/infrared imaging spectrometer (AVIRIS) sensor over Salinas Valley, Southern California, in 1998. The image size is $512 \times 217$ pixels with a spatial resolution of 3.7 m. It contains 224 bands and 16 ground-truth classes in total. After removing 20 bands that are severe affected by noise, the remaining 204 bands are used for the experiments.The Salinas scene in false color and its corresponding ground truth are shown in Fig. 5.
\begin{figure}[!htbp]
\centering
\includegraphics[scale=0.41]{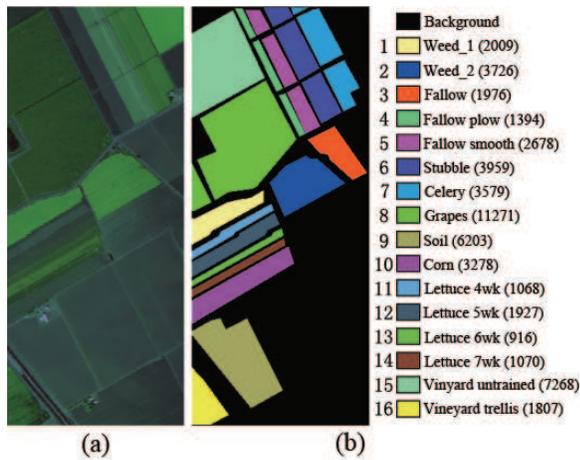}
\caption{Salinas hyperspectral remote sensing image. (a) HSI in false color. (b) Corresponding ground truth. (Note that the number of samples for each class is shown in brackets.)}
\label{Salinas_gt_fig}
\end{figure}
\begin{figure*}[!htbp]
\centering
\includegraphics[scale=0.85]{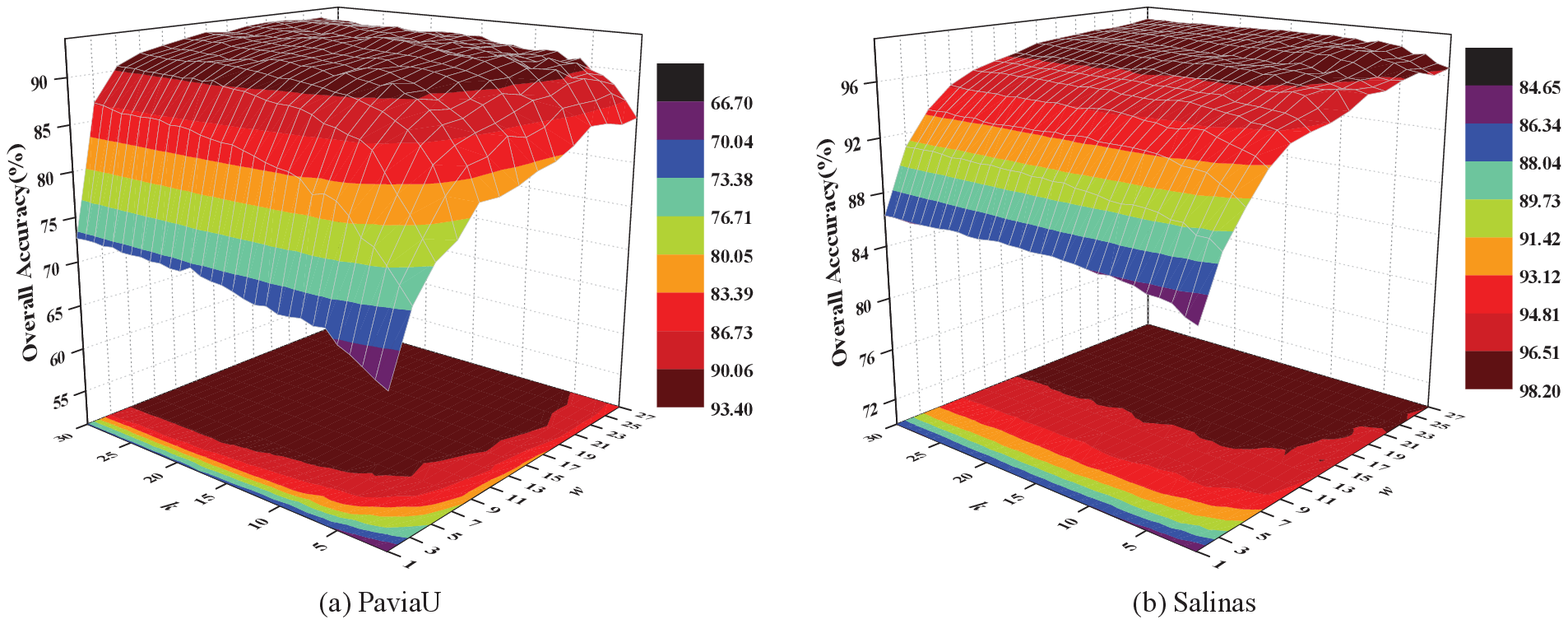}
\caption{Classification accuracies for PaviaU and Salinas data sets under different \emph{w} and \emph{k}.}
\label{PaviaU_Salinas_w_and_k_fig}
\end{figure*}
\subsection{Experimental Setup}
In each experiment, the HSI data set is randomly divided into training and test sets. The training set is used to learn a low-dimensional embedding space using different DR methods. Then, all test samples are projected onto the embedding space. After that, the nearest neighbor classifier (1-NN) is employed for classification in all experiments. Overall classification accuracies (OAs), average classification accuracies (AAs), and the \emph{kappa} coefficients ($\kappa$) are used to evaluate the classification performance. To robustly evaluate the results in different training conditions, we randomly chose training samples and repeated the experiment five times in each condition.

The proposed SSMRPE method was compared with several state-of-art DR algorithms such as PCA, NPE, LPP, LDA, LFDA, SC-NPE, DSSMs and LPNPE. The former five methods are spectral-based DR methods, while the later three approaches make use of spatial relationship and spectral information for DR of HSI data.

In order to evaluate the classification influence of spatial window size \emph{w} and neighbor number \emph{k}, we randomly selected 1\% samples from the two HSI datasets for training, and the remaining samples are used for testing. Parameters \emph{w} and \emph{k} were tuned with a set of $\left\{ 1, 3, 5, \cdots, 25, 27 \right\}$ and a set of $\left\{ 1, 2, 3, \cdots, 59, 60 \right\}$, respectively. Fig. 6 shows the OAs versus different values of \emph{w} and \emph{k}.

As can be seen from Fig. 6, the performance of the SSMRPE becomes much better with large values of parameters \emph{k} and \emph{w}. The OAs of SSMRPE first improve with the increase of \emph{w}, and then the OAs maintained a stable value. The reason is that a larger spatial window contains more neighbors, which brings benefits to extract discriminant features for classification. However, if the size of spatial window is too large, the spatial information in neighbors will become redundant for DR of HSI. Furthermore, a larger spatial window also leads to a great increase in the computational complexity. Based on the above analysis, we choose \emph{w} = 13, \emph{k} = 20 for PaviaU dataset, and \emph{w} = 15, \emph{k} = 15 for Salinas dataset in the following experiments.

\subsection{Classification Results}
In this section, the experiments were conducted on PaviaU and Salinas data sets to evaluate the classification performance of different DR methods. In order to demonstrate the classification performance of different DR algorithms under different training conditions, we randomly selected $n_i$ ($n_i$ =20, 30, 40, and 50) samples from each class for training, and the others for testing. Table I is the classification results on PaviaU dataset under different number of training samples.
\begin{table}[!htbp]
\centering
\footnotesize
\setlength{\tabcolsep}{1.5mm}
\renewcommand{\arraystretch}{1.6}
\caption{Classification results using different DR methods with KNN for the PaviaU data set. [OA $\pm$ std(\%)]}
\label{PaviaU_Classifiers}
\begin{tabular}{cccccc}
\hline
Algorithm & $n_i=20$           & $n_i=30$           & $n_i=40$           & $n_i=50$          \\
\hline
RAW       & 74.53 $\pm$ 1.24   & 78.05 $\pm$ 0.42   & 81.40 $\pm$ 0.75   & 84.32 $\pm$ 0.76  \\
PCA       & 74.53 $\pm$ 1.25   & 78.05 $\pm$ 0.41   & 81.39 $\pm$ 0.75   & 84.30 $\pm$ 0.78  \\
NPE       & 72.12 $\pm$ 1.42   & 80.57 $\pm$ 1.11   & 86.60 $\pm$ 0.85   & 89.24 $\pm$ 1.07  \\
LPP       & 71.37 $\pm$ 1.44   & 81.12 $\pm$ 0.75   & 86.97 $\pm$ 1.59   & 90.03 $\pm$ 0.62   \\
LDA       & 83.48 $\pm$ 1.14   & 91.70 $\pm$ 0.30   & 92.96 $\pm$ 1.20   & 93.28 $\pm$ 0.73  \\
LFDA      & 82.37 $\pm$ 1.17   & 90.95 $\pm$ 0.85   & 93.05 $\pm$ 1.28   & 94.13 $\pm$ 0.88  \\
DSSMs     & 74.36 $\pm$ 1.31   & 79.97 $\pm$ 1.32   & 83.39 $\pm$ 1.36   & 85.78 $\pm$ 0.60  \\
SC-NPE    & 72.52 $\pm$ 1.45   & 80.27 $\pm$ 1.27   & 82.96 $\pm$ 1.23   & 85.82 $\pm$ 0.94   \\
LPNPE     & 85.77 $\pm$ 1.27   & 90.28 $\pm$ 1.25   & 93.76 $\pm$ 1.25   & 95.37 $\pm$ 1.04  \\
SSMRPE    & \textbf{86.92} $\pm$ 1.22   & \textbf{92.31} $\pm$ 1.33   & \textbf{95.78} $\pm$ 1.29   & \textbf{97.93} $\pm$ 0.59   \\
\hline
\end{tabular}
\end{table}

From Table I, we can see that the classification accuracies of all DR algorithms improved with the increase of the number of training samples. The reason is that a large number of training samples contain more available information to learn low-dimensional embedding features. The supervised spectral-based methods, LDA and LFDA, are superior to unsupervised ones such as PCA, NPE, and LPP, which indicates the prior knowledge of training samples is useful for DR of HSI data. For spatial-spectral combined methods, the proposed SSMRPE method achieves better classification results than DSSMs, SC-NPE, and LPNPE in most conditions. This is because SSMRPE explores a new spatial-spectral combined distance to choose effective neighbors which are used to construct a spatial-spectral adjacency graph for discovering the intrinsic manifold structure of HSI data, and the reconstruction weights of spatial neighbors are adjusted to enhance the aggregation of HSI data. Thus, the discriminating power of extracted features is further improved.
\begin{table*}[htbp]
\centering
\footnotesize
\setlength{\tabcolsep}{3mm}
\renewcommand{\arraystretch}{1.6}
\caption{CLASSIFICATION ACCURACIES (\%) FOR THE PAVIAU DATA SET (1\% LABELED SAMPLES PER CLASS FOR TRAINING, TOTALLY 427 TRAINING SAMPLES AND 42349 TEST SAMPLES) (\%)}
\begin{tabular}{ccccccccccccc}
\hline
Class & Train & Test  & RAW   & PCA   & LDA   & LFDA  & NPE   & LPP   & SCNPE & DSSMs & LPNPE & SSMRPE \\
\hline
1 & 66 & 6565     & 92.14   & 92.14   & 92.93   & 96.31   & 94.96   & 96.53   & 93.71   & 92.22   & 94.36   & \textbf{97.62} \\
2 & 186 & 18463   & 96.65   & 96.65   & 99.60   & 99.06   & 98.03   & 99.02   & 98.46   & 96.86   & \textbf{99.87}   & 99.71 \\
3 & 21 & 2078     & 85.37   & 85.37   & 94.71   & 91.00   & 90.95   & 91.67   & 84.46   & 87.42   & 93.55   & \textbf{95.24} \\
4 & 31 & 3033     & 76.69   & 76.69   & \textbf{92.19}   & 91.20   & 78.44   & 86.71   & 78.96   & 76.66   & 83.55   & 87.67 \\
5 & 13 & 1332     & 99.77   & 99.77   & 99.32   & 99.70   & 99.85   & 99.85   & \textbf{99.92}   & 99.77   & 99.77   & \textbf{99.92} \\
6 & 50 & 4979     & 74.01   & 74.01   & 95.90   & 96.85   & 90.88   & 96.49   & 86.46   & 73.99   & \textbf{100.00}  & 99.68 \\
7 & 13 & 1317     & 73.04   & 73.04   & 99.70   & 97.27   & 82.69   & 92.26   & 87.17   & 72.67   & 91.95   & \textbf{99.85} \\
8 & 37 & 3645     & 87.76   & 87.76   & 84.01   & 72.65   & \textbf{92.15}   & 85.54   & 87.11   & 87.94   & 84.94   & 89.22 \\
9 & 10 & 937      & \textbf{95.20}   & \textbf{95.20}   & 91.14   & 89.97   & 94.66   & 94.56   & 94.13   & \textbf{95.20}   & 92.53   & 93.92 \\
\hline
OA & - & - & 89.87 & 89.87 & 95.82 & 94.90 & 93.96 & 95.65 & 92.85 & 90.68 & 95.94 & \textbf{97.27} \\
AA & - & - & 86.73 & 86.73 & 94.38 & 92.66 & 91.40 & 93.62 & 90.04 & 87.51 & 94.36 & \textbf{95.86} \\
Kappa & - & - & 86.39 & 86.39 & 94.46 & 93.24 & 91.94 & 94.22 & 90.43 & 87.22 & 94.49 & \textbf{96.39}  \\
\hline
\end{tabular}
\end{table*}
\par In order to analyze the classification performances of different DR algorithms on each class, 1\% samples per class were randomly selected for training and the rest for testing. The classification accuracy of each class, OA, AA and \emph{kappa} coefficient in PaviaU data set were shown in Table II. Fig. 7 is the corresponding classification maps of different methods for PaviaU hyperspectral image.
\begin{figure*}[!htbp]
\centering
\includegraphics[scale=0.6]{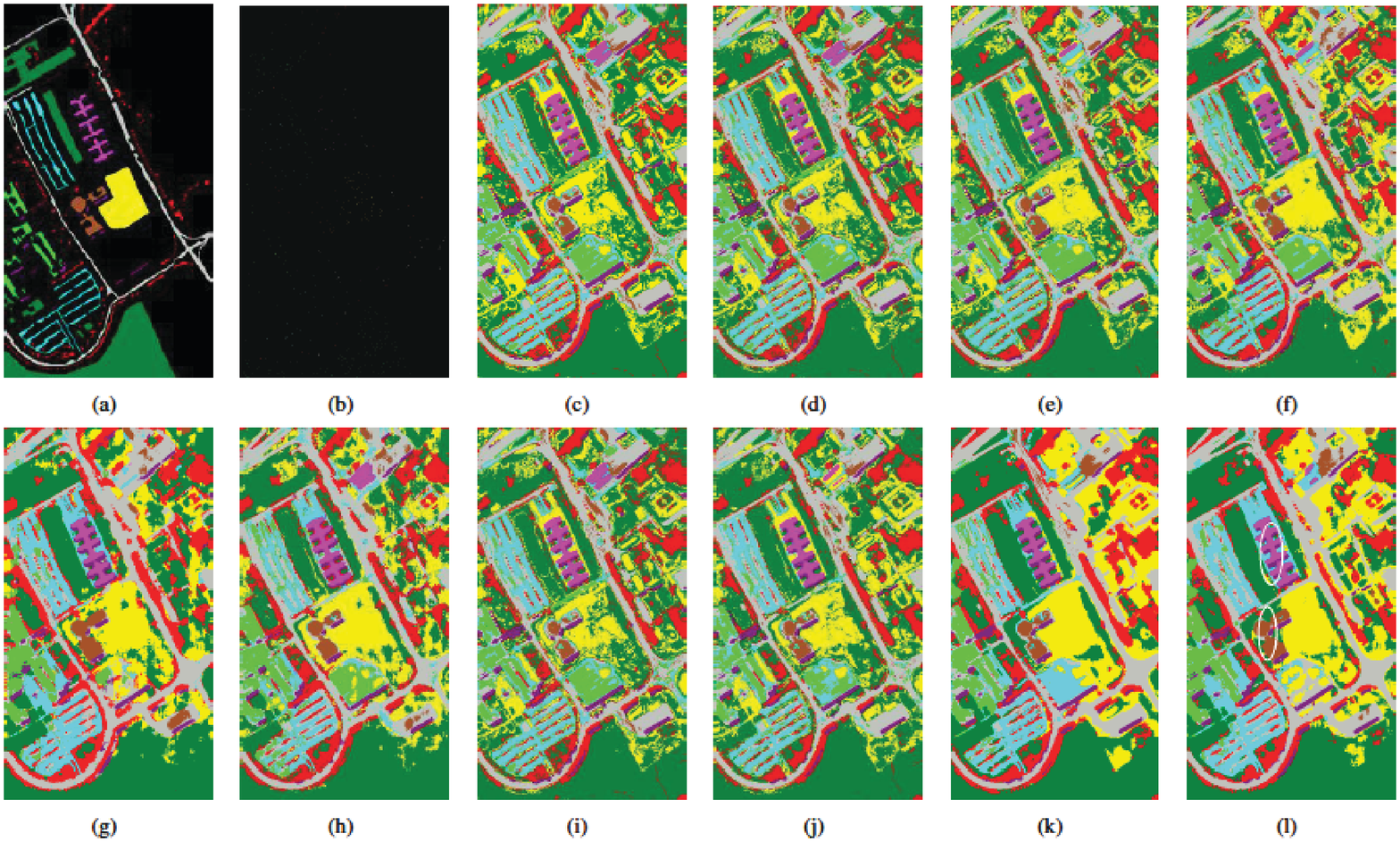}
\caption{Classification maps for different methods based on KNN for the PaviaU data set. (a)  Ground truth (b) Training samples (c) RAW (d) PCA (e) NPE (f) LPP (g) LDA (h) LFDA (i) SCNPE (j) DSSMs (k) LPNPE (l) SSMRPE}
\label{PaviaU_map}
\end{figure*}
\par As shown in Table II, the unsupervised spatial-spectral combined methods, SC-NPE, LPNPE and SSMRPE, perform better than unsupervised spectral-based methods such as RAW, PCA, NPE and LPP, which indicates that combining spatial structure and spectral information provided by training samples gives the benefits to DR of HSI. Our proposed method obtained the strongest classification effect in most classes and achieved the best OA, AA, and \emph{kappa} coefficient. The reason is that SSMRPE constructs a spatial-spectral adjacency graph to reveal the intrinsic structure of HSI data, and it also exploits the spatial distance to adjust the reconstruction weights between pixels and their neighbors for enhancing discriminating ability of embedding features. The numerical results shown in Table II are confirmed by inspecting the classification maps in Fig. 7, where smoother classification map is obtained by SSMRPE.
\par Table III reports the classification performances of different DR algorithms on Salinas hyperspectral data set under different number of training samples. It is obviously that the overall accuracy of each method improved with the increasing number of training samples. The proposed algorithm achieved the best classification performance than other DR methods in most cases. The classification results indicate that SSMRPE not only inherits the merit of manifold learning but also makes full use of the spatial consistency of HSI data, which is beneficial to extract discriminating features and subsequently enhance classification performance.
\begin{table}[htbp]
\centering
\footnotesize
\setlength{\tabcolsep}{1.5mm}
\renewcommand{\arraystretch}{1.6}
\caption{Classification results using different DR methods with KNN for the Salinas data set. [OA $\pm$ std(\%)]}
\label{PaviaU_Classifiers}
\begin{tabular}{cccccc}
\hline
Algorithm & $n_i=20$           & $n_i=30$           & $n_i=40$           & $n_i=50$   \\
\hline
RAW       & 90.46 $\pm$ 1.04 & 91.96 $\pm$ 0.74 & 92.93 $\pm$ 1.62 & 93.31 $\pm$ 0.36  \\
PCA       & 90.45 $\pm$ 1.06 & 91.95 $\pm$ 0.77 & 92.92 $\pm$ 1.66 & 93.31 $\pm$ 0.36  \\
NPE       & 91.37 $\pm$ 1.20 & 92.95 $\pm$ 1.33 & 93.06 $\pm$ 1.90 & 94.02 $\pm$ 0.71 \\
LPP       & 90.57 $\pm$ 0.46 & 94.77 $\pm$ 1.06 & 95.42 $\pm$ 1.72 & 96.47 $\pm$ 0.63 \\
LDA       & 95.27 $\pm$ 0.81 & 96.19 $\pm$ 1.30 & 96.68 $\pm$ 1.16 & 97.39 $\pm$ 0.70 \\
LFDA      & 89.54 $\pm$ 0.71 & 93.71 $\pm$ 0.91 & 95.89 $\pm$ 1.37 & 96.22 $\pm$ 0.42 \\
DSSMs     & 90.43 $\pm$ 0.65 & 91.96 $\pm$ 0.94 & 93.23 $\pm$ 0.81 & 94.33 $\pm$ 0.41 \\
SC-NPE    & 91.23 $\pm$ 0.51 & 93.12 $\pm$ 0.93 & 93.78 $\pm$ 0.74 & 95.20 $\pm$ 0.39 \\
LPNPE     & 94.51 $\pm$ 0.64 & 96.08 $\pm$ 0.92 & 96.86 $\pm$ 0.69 & 97.70 $\pm$ 0.81 \\
SSMRPE    & \textbf{95.82} $\pm$ 1.03 & \textbf{97.37} $\pm$ 0.56 & \textbf{97.94} $\pm$ 1.15 & \textbf{99.23} $\pm$ 0.73 \\
\hline
\end{tabular}
\end{table}
\begin{table*}[htbp]
\centering
\footnotesize
\setlength{\tabcolsep}{3mm}
\renewcommand{\arraystretch}{1.6}
\caption{CLASSIFICATION ACCURACIES (\%) FOR THE Salinas DATA SET (1\% LABELED SAMPLES PER CLASS FOR TRAINING, TOTALLY 544 TRAINING SAMPLES AND 53585 TESTING SAMPLES) (\%)}
\begin{tabular}{ccccccccccccc}
\hline
Class & Train & Test & RAW & PCA & LDA & LFDA & NPE & LPP & SCNPE & DSSMs & LPNPE & SSMRPE \\
\hline
1 & 20 & 1989    & 99.50    & 99.50   & \textbf{100.00}   & 99.85   & 99.70   & 99.95   & 99.80   & 98.74   & 99.85   & \textbf{100.00} \\
2 & 37 & 3689    & 99.54    & 99.54   & \textbf{100.00}   & \textbf{100.00 } & 99.73   & \textbf{100.00}  & 99.76   & 99.76   & 99.97   & \textbf{100.00} \\
3 & 20 & 1956    & 89.98    & 89.98   & \textbf{100.00}   & 99.59   & \textbf{100.00}  & \textbf{100.00}  & \textbf{100.00}  & 95.81   & \textbf{100.00}  & \textbf{100.00} \\
4 & 14 & 1380    & 98.99    & 98.99   & \textbf{99.78}    & 99.49   & 99.20   & 99.71   & 98.91   & 98.99   & 98.77   & 99.20 \\
5 & 27 & 2651    & 96.27    & 96.27   & 99.02    & \textbf{99.06}   & 97.62   & 98.87   & 98.49   & 95.13   & 99.02   & 98.87 \\
6 & 40 & 3919    & 99.82    & 99.82   & 99.92    & 99.87   & 99.85   & 99.92   & 99.87   & 99.80   & 99.87   & \textbf{99.97} \\
7 & 36 & 3543    & 98.62    & 98.62   & \textbf{99.97}    & 99.92   & 99.32   & \textbf{99.97 }  & 98.96   & 99.07   & \textbf{99.97}   & \textbf{99.97} \\
8 & 113 & 11158  & 85.30    & 85.30   & 96.07    & 89.28   & 91.57   & 93.04   & 89.83   & 86.54   & 95.90   & \textbf{98.37} \\
9 & 62 & 6141    & 98.47    & 98.47   & \textbf{100.00}   & \textbf{100.00 } & \textbf{100.00}  & 99.95   & \textbf{100.00}  & 98.84   & 99.98   & \textbf{100.00} \\
10 & 33 & 3245   & 93.90    & 93.90   & \textbf{99.69 }   & 99.29   & 95.10   & 96.39   & 96.18   & 94.05   & 97.16   & 98.37 \\
11 & 11 & 1057   & 80.79    & 80.79   & 99.43    & 99.72   & 98.86   & 99.15   & \textbf{100.00}  & 95.84   & 99.91   & \textbf{100.00} \\
12 & 19 & 1908   & 99.90    & 99.90   & \textbf{100.00}   & \textbf{100.00}  & 99.95   & \textbf{100.00}  & \textbf{100.00}  & 99.84   & \textbf{100.00}  & \textbf{100.00} \\
13 & 10 & 906    & 97.46    & 97.46   & \textbf{99.01}    & 98.79   & 98.01   & 98.45   & 98.01   & 97.35   & 98.12   & \textbf{99.01} \\
14 & 11 & 1059   & 93.39    & 93.39   & 97.26    & 94.62   & 93.86   & 96.69   & 93.77   & 97.17   & 97.36   & \textbf{98.11} \\
15 & 73 & 7195   & 87.06    & 87.06   & 90.33    & 83.64   & 87.46   & 91.87   & 84.38   & 86.84   & 97.32   & \textbf{98.42} \\
16 & 18 & 1789   & 97.21    & 97.21   & \textbf{99.83 }   & 99.16   & 99.22   & 99.78   & 94.13   & 96.26   & 99.50   & \textbf{99.83} \\
\hline
OA & - & - & 93.27 & 93.27 & 97.71 & 95.27 & 95.83 & 94.99 & 94.50 & 94.56 & 98.41 & \textbf{99.28 }\\
AA & - & - & 94.76 & 94.76 & 98.76 & 97.64 & 97.46 & 97.02 & 97.36 & 96.25 & 98.91 & \textbf{99.37} \\
Kappa & - & - & 92.52 & 92.52 & 97.45 & 94.74 & 95.36 & 94.43 & 93.87 & 93.40 & 98.23 & \textbf{99.12}\\
\hline
\end{tabular}
\end{table*}
\par In order to evaluate the performance of the SSMRPE method under different training conditions, we use a balanced training set in which around 1\% of the labeled samples per class have been randomly selected for training and the remaining samples are used for testing. Table IV reports the classification accuracies of different DR methods in the Salinas data set, and Fig. 8 shows the corresponding classification maps.
\begin{figure*}[!htbp]
\centering
\includegraphics[scale=0.9]{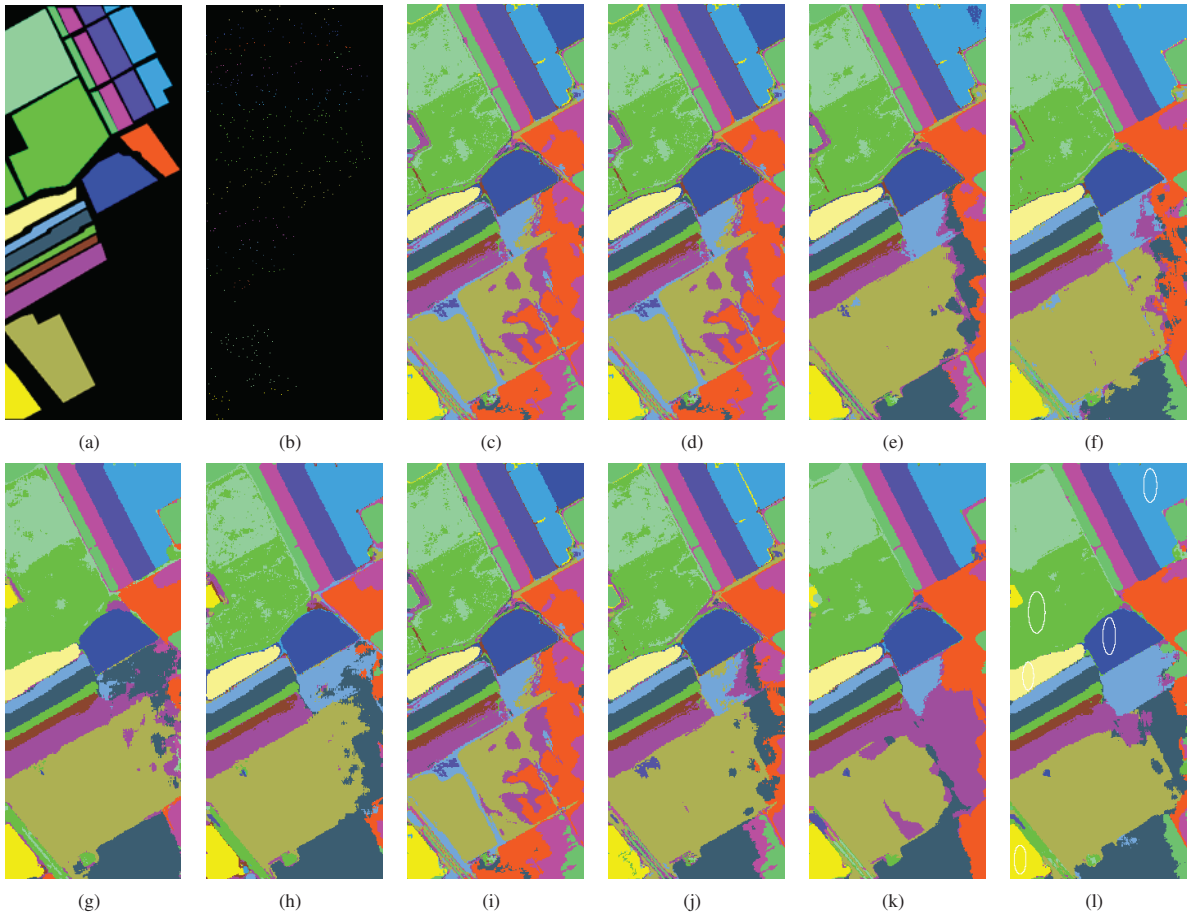}
\caption{Classification maps for different methods based on KNN for the Salinas data set. (a)  Ground truth (b) Training samples (c) RAW (d) PCA (e) NPE (f) LPP (g) LDA (h) LFDA (i) SCNPE (j) DSSMs (k) LPNPE (l) SSMRPE}
\label{Salinas_map}
\end{figure*}

In Table IV, the proposed SSMRPE method obtained better classification accuracy in most classes, and achieved the best results in terms of OA, AA, and \emph{kappa} coefficient among all DR methods with balanced training data. As shown in Fig. 8, the numerical results are confirmed by visual inspection of the classification maps. The SSMRPE method produces more homogenous areas and smoother classification maps than the other methods, especially in the extreme case (e.g., classes ${\rm{Weeds\_1, Weeds\_2, lettuce\_4wk, lettuce\_6wk}}$). The probable reason is that it not only explores the spatial-spectral adjacency graph to reveal the intrinsic manifold structure of HSI data but also utilizes the spatial distance to adjust the reconstruction weights for extracting the discriminant features in the low-dimensional embedding space, which is more beneficial to HSI classification.

\section{Conclusion}
In this paper, we proposed an unsupervised DR method termed spatial-spectral manifold reconstruction preserving embedding (SSMRPE) to learn the low-dimensional embedding features for HSI classification. SSMRPE proposed a new spatial-spectral combined distance to construct the spatial-spectral adjacency graph for revealing intrinsic manifold structure of HSI data. Then, it adjusts the reconstruction weights according to the spatial relationship between each point and its neighbors to improve the efficiency of manifold reconstruction. As a result, the proposed method can effectively extract the discriminant features and subsequently improve HSI classification performance. Experimental results on PaviaU and Salinas hyperspectral datasets show that the proposed algorithm performs much better than some state-of-the-art DR methods in terms of classification accuracy and \emph{kappa} coefficient. Our future work will focus on how to reduce the computational complexity and further improve the classification performance  of the SSMRPE method.

\end{document}